\documentclass[11pt]{article}

\usepackage[final]{acl}

\usepackage{times}
\usepackage{latexsym}

\usepackage[T1]{fontenc}

\usepackage[utf8]{inputenc}

\usepackage{microtype}

\usepackage{inconsolata}

\usepackage{graphicx}

\usepackage{booktabs}
\usepackage{amsmath}
\usepackage{xspace}
\usepackage{ulem}
\usepackage{subcaption}
\usepackage[most]{tcolorbox}
\usepackage{amssymb}
\usepackage{soul}
\usepackage[table,dvipsnames]{xcolor}
\definecolor{mygreen}{RGB}{19,163,157}
\definecolor{myred}{RGB}{241,82,44}

\newenvironment{itemize*}%
 {\leftmargini=10pt\begin{itemize}%
  \setlength{\itemsep}{0pt}%
  \setlength{\parskip}{0pt}%
  }%
 {\end{itemize}}
\newenvironment{enumerate*}%
 {\leftmargini=10pt\begin{enumerate}%
  \setlength{\itemsep}{0pt}%
  \setlength{\parskip}{0pt}}%
 {\end{enumerate}}

\newcommand{\highlight}[2]{\sethlcolor{#1}\hl{#2}}
\definecolor{cyan10}{HTML}{E5F6FF}
\definecolor{cyan20}{HTML}{BAE6FF}
\definecolor{cyan60}{HTML}{0072c3}
\definecolor{cyan70}{HTML}{00539a}
\definecolor{cyan80}{HTML}{003a6d}
\definecolor{teal10}{HTML}{D9FBFB}
\definecolor{teal20}{HTML}{9EF0F0}
\definecolor{teal60}{HTML}{007d79}
\definecolor{orange10}{HTML}{FFF2E8}
\definecolor{orange20}{HTML}{FFD9BE}
\definecolor{orange60}{HTML}{ba4e00}
\definecolor{blue10}{HTML}{EDF5FF}
\definecolor{blue20}{HTML}{D0E2FF}
\definecolor{blue70}{HTML}{0043ce}
\definecolor{blue80}{HTML}{002d9c}
\definecolor{magenta10}{HTML}{FFF0F7}
\definecolor{magenta20}{HTML}{FFD6E8}
\definecolor{magenta30}{HTML}{ffafd2}
\definecolor{magenta50}{HTML}{ee5396}
\definecolor{magenta60}{HTML}{d02670}
\definecolor{magenta70}{HTML}{9f1853}
\definecolor{purple10}{HTML}{F6F2FF}
\definecolor{purple20}{HTML}{E8DAFF}
\definecolor{purple30}{HTML}{d4bbff}
\definecolor{purple70}{HTML}{8a3ffc}
\definecolor{rose10}{HTML}{FCF2ED}
\definecolor{rose20}{HTML}{F9D9D1}
\definecolor{rose60}{HTML}{ab5638}
\definecolor{rose70}{HTML}{853c27}
\definecolor{red10}{HTML}{FFF1F1}
\definecolor{red20}{HTML}{FFD7D9}
\definecolor{green10}{HTML}{DEFBE6}
\definecolor{green20}{HTML}{A7F0BA}
\definecolor{green70}{HTML}{0e6027}
\definecolor{green80}{HTML}{044317}
\definecolor{yellow10}{HTML}{fcf4d6}
\definecolor{yellow20}{HTML}{fddc69}
\definecolor{gray20}{HTML}{e0e0e0}
\definecolor{gray30}{HTML}{c6c6c6}
\definecolor{gray40}{HTML}{a8a8a8}
\definecolor{gray80}{HTML}{393939}

\newcommand{\ie}{\textit{i.e.}\xspace} 
\newcommand{\eg}{\textit{e.g.}\xspace} 

\title{When LLMs Read Tables Carelessly: \\ Measuring and Reducing Data Referencing Errors}

\author{
  \textbf{Yuqing Yang\textsuperscript{1}}\thanks{Work done during an internship at Amazon. Email: \href{mailto:yyang063@usc.edu}{yyang063@usc.edu}.},
  \textbf{Qi Zhu\textsuperscript{2}}\thanks{Corresponding author. Email: \href{mailti:qzhuamzn@amazon.com}{qzhuamzn@amazon.com}.},
  \textbf{Zhen Han\textsuperscript{2}},
  \textbf{Boran Han\textsuperscript{2}},
\\
  \textbf{Zhengyuan Shen\textsuperscript{2}},
  \textbf{Shuai Wang\textsuperscript{2}},
  \textbf{Vassilis N. Ioannidis\textsuperscript{2}},
  \textbf{Huzefa Rangwala\textsuperscript{2}}
\\
\\
  \textsuperscript{1}University of Southern California,
  \textsuperscript{2}AWS AI Labs
}

\begin{document}
\maketitle
\begin{abstract}

While large language models (LLMs) perform well on table tasks, they still make data referencing errors (DREs), \ie, incorrectly citing or omitting table values, despite understanding the table structure. Beyond final-answer accuracy, DREs directly compromise the correctness and reliability of intermediate reasoning steps. Yet prior studies have only offered limited, small-scale analyses. In this work, we present the first systematic evaluation of tabular data referencing errors across different models and tasks. Our results show that DREs occur across all tested models (1.7B to 20B parameters). Furthermore, we demonstrate that incorporating data referencing as a critic significantly improves answer accuracy up to 12.0\%, through critic-based filtering and rejection sampling. Finally, we trained a lightweight 4B-parameter critic model that achieves an average F1 score of 78.2\% in detecting both in-distribution and out-of-distribution DREs, and effectively assists inference for larger models.

\end{abstract}

\section{Introduction}
\begin{figure}[t]
    \centering
    \begin{subfigure}{\linewidth}
        \centering
        \includegraphics[width=0.95\linewidth]{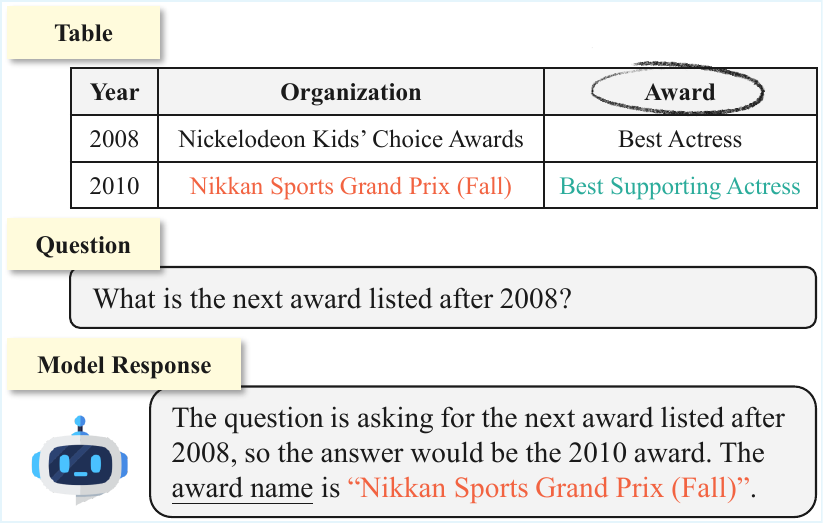}
        \caption{\small Incorrect Citation: Model confuses the ``\textcolor{myred}{Organization}'' column with the ``\textcolor{mygreen}{Award}'' column.}
        \label{fig:intro-1}
    \end{subfigure}
    \begin{subfigure}{\linewidth}
        \centering
        \includegraphics[width=0.95\linewidth]{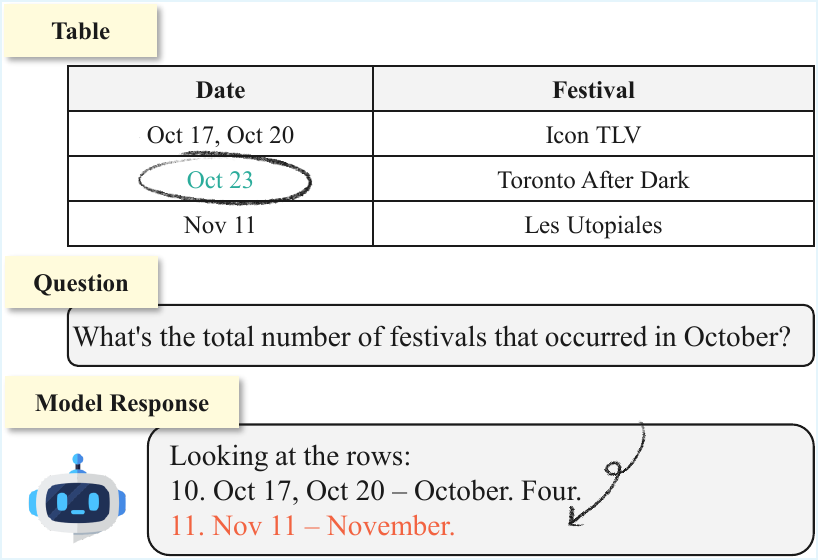}
        \caption{\small Omitted Information: Model omits the row of ``\textcolor{mygreen}{Oct 23}''.}
        \vspace{-1mm}
        \label{fig:intro-2}
    \end{subfigure}
    \caption{Illustration of Tabular DREs.}
    \label{fig:intro}
    \vspace{-4mm}
\end{figure}

Tables are one of the most common ways to represent information, providing a structured format for organizing data. They are widely used across real-world domains such as finance \citep{DBLP:conf/emnlp/ChenCSSBLMBHRW21}, healthcare \citep{DBLP:conf/iclr/YanCH0H0025}, and scientific reporting \citep{DBLP:conf/nips/MoosaviRRG21,DBLP:conf/acl/ZhangWWDLXWZ25}, making the ability to effectively perform tasks over tabular data essential. Solving table-related tasks requires several capabilities: understanding tables presented in textual formats, accurately locating and citing relevant values, and reasoning over critical values to derive correct answers. Large Language Models (LLMs) are increasingly applied to these tasks and often achieve strong performance \citep{DBLP:journals/corr/abs-2505-23621,DBLP:journals/corr/abs-2505-12415,DBLP:journals/corr/abs-2506-01710}, yet they still commit surprisingly basic mistakes even when the table format is correctly parsed, by referencing table content incorrectly, as illustrated in Figure~\ref{fig:intro}.

A primary source of such errors is the dense and similar structure of tables, which makes it hard for models to reliably locate and cite values. For example, answering ``Which country had the highest GDP growth between 2020 and 2022?'' requires aligning multiple year columns across rows, where, analogous to human oversight, a slip can lead to mistakes. We refer to such failures to faithfully retrieve and cite information from the input as \textbf{\emph{Data Referencing Errors}} (DREs). These errors can degrade response quality and sometimes final-answer accuracy, yet they are not fully captured by final accuracy metrics alone. Although prior work \citep{DBLP:journals/corr/abs-2505-15110,DBLP:journals/corr/abs-2501-19378} has observed DREs, analyses remain narrow, typically limited to a single model and a small set of human-annotated cases. In this work, we systematically investigate how prevalent DREs are, how they can be effectively mitigated, and how mitigating them influences final-answer accuracy.

We first categorize tabular DREs into two types: \textbf{Incorrect Citation}, involving individual values, and \textbf{Omitted Information}, involving entire relevant portions, as illustrated in Figure~\ref{fig:intro}. We then employ LLM-as-a-Judge framework \citep{DBLP:conf/nips/ZhengC00WZL0LXZ23} to automatically detect DREs given a table and a generation model's response. Upon evaluation, We find that DREs are ubiquitous across different models (from 1.7B to 20B parameters) and across diverse table-related tasks (including Question Answering, Claim Verification, and Table-to-Text). They are not effectively eliminated by either reasoning models' self-reflection mechanisms \citep{DBLP:journals/corr/abs-2408-03314,DBLP:journals/corr/abs-2501-19393,DBLP:journals/corr/abs-2502-03387} or by prompting-based approaches. For instance, Qwen3-8B \citep{DBLP:journals/corr/abs-2505-09388}, even with extended self-reflection, exhibits a 14.04\% DRE rate (i.e., the proportion of responses containing DREs) on the WTQ \citep{DBLP:conf/acl/PasupatL15} dataset, and still 12.50\% when further prompted not to miscite or omit table content.

Incorporating DRE detection as a critic not only improves the quality of intermediate reasoning steps beyond what is captured by final-answer accuracy rewards, but also noticeably enhances overall performance. We explore two approaches. First, \textbf{critic-based filtering}, which selects the subset of sampled responses with minimal DREs, yields substantially higher accuracy than using all sampled responses, and can further enhance majority voting when combined. Second, \textbf{rejection sampling}, which repeatedly resamples response segments until the critic accepts it, obtains consistent gains and can improve accuracy by up to 11.96\%. Notably, DREs are largely avoidable rather than fundamental limitations, yet rejection sampling with the critic offers a more robust way to reduce their occurrence.

Finally, given the high cost and black-box nature of using larger models as critics, we investigate the potential of small-scale LLMs (\ie, Qwen3-4B-Instruct, \citealp{DBLP:journals/corr/abs-2505-09388}) for detecting DREs. To this end, we construct training data from Qwen3-8B responses on the WTQ training set and adopt a two-stage training procedure: supervised fine-tuning for warm-up, followed by RLVR (reinforcement learning with verified reward, \citealp{DBLP:journals/corr/abs-2411-15124,DBLP:journals/corr/abs-2501-12948}) to enhance robustness in DRE detection without constraining the Chain-of-Thought (CoT) format. Our experiments show that the trained Critic-4B consistently outperforms the untrained baseline both in-distribution and out-of-distribution, achieving an average improvement of 8.65\% F1. Moreover, we demonstrate that this lightweight critic can mitigate DREs across different models and improve final accuracy more effectively than prompting-based methods.\footnote{Our code is available at \url{https://github.com/ayyyq/table-referencing}.}

Our work shed light on the overlooked issue of data referencing errors, a unique error pattern that, while particularly common in table-related tasks, also appears in other domains. Although these broader cases are beyond the scope of this paper, we aim to inspire follow-up work to improve both critic and generation models by enhancing their data referencing capabilities.

\section{Related Work}
\paragraph{Table LLMs}
Solving tabular tasks using LMs has been a long-standing research topic. Early work relied on table pre-training with specialized architectures, such as TaPas \citep{DBLP:conf/acl/HerzigNMPE20} and TaBERT \citep{DBLP:conf/acl/YinNYR20}. With the scaling of general-purpose LLMs, recent methods adapt them to tabular settings through prompting \citep{DBLP:conf/sigir/YeHYLHL23,DBLP:conf/emnlp/JiangZDYZW23}, supervised fine-tuning \citep{DBLP:conf/naacl/ZhangYL024,DBLP:journals/corr/abs-2411-02059,DBLP:conf/acl/ZhangLZMZ0L0XZZ25}, and reinforcement learning \citep{DBLP:journals/corr/abs-2505-23621,DBLP:journals/corr/abs-2505-12415,DBLP:journals/corr/abs-2506-01710}. As LLMs become increasingly powerful, particularly with the emergence of reasoning models that demonstrate strong problem-solving capabilities through extended thinking processes \citep{DBLP:journals/corr/abs-2501-12948,DBLP:journals/corr/abs-2505-09388,openai2025gptoss120bgptoss20bmodel}, they can already exhibit strong baseline performance on table tasks without task-specific training, as evidenced by \citet{DBLP:journals/corr/abs-2505-23621,DBLP:journals/corr/abs-2505-12415} and our experiments in Table~\ref{tab:dre}. These developments call for moving beyond final-answer accuracy toward more fine-grained limitations.

\paragraph{Evaluation Beyond Accuracy}

Most evaluation benchmarks emphasize final accuracy for simplicity, overlooking the quality of intermediate reasoning. To address this, prior work has proposed \textit{process-level reward models} (PRMs, \citealp{DBLP:conf/iclr/LightmanKBEBLLS24,DBLP:conf/acl/ZhangZWZLYLZL25}), which evaluate reasoning steps rather than only outcomes. Related efforts further decompose evaluation into dimensions: e.g., validity and redundancy in mathematical reasoning \citep{DBLP:conf/aaai/Xia0LWL25}, instruction-following and truthfulness in alignment \citep{DBLP:conf/icml/CuiY0YH0NXXL0024}, and relevance and completeness in long-form QA \citep{DBLP:conf/nips/WuHSDSASOH23}. In table reasoning, however, benchmarks still focus almost exclusively on final correctness \citep{DBLP:conf/aaai/WuYCZLDLSCSLLN25,DBLP:conf/acl/PasupatL15}. We introduce data referencing errors as a complementary dimension that captures how reliably models use table values, reflecting both intermediate reasoning quality and final performance.

\paragraph{Existing Work on DREs}
Table-related tasks, especially Table QA, require models to use table values both completely and accurately. Prior work has recognized this need. For example, \citet{DBLP:journals/corr/abs-2505-15110} analyzed 50 WTQ samples from Distill-Llama-8B \citep{DBLP:journals/corr/abs-2501-12948} and found that more than 80\% of errors came from incorrect locating and citation. Yet such studies do not systematically characterize DREs. Other work \citep{DBLP:journals/corr/abs-2505-12415,DBLP:journals/corr/abs-2506-01710} introduces auxiliary rewards to improve table referencing, but relies on supervised fine-tuning with annotated table regions. For instance, models are trained to generate special tags such as {\small \texttt{<|cell content|><|column name|>}} when using the specific table values \textit{needed} to answer a question. In contrast, we evaluate models' overall accuracy in referencing \textit{any} table values and analyze how DREs affect performance. Our framework further supports critic-based detection that can be seamlessly integrated into existing LLMs, improving both response quality and final-answer accuracy without requiring special annotations or disrupting reasoning chains \citep{tang2025eigen1adaptivemultiagentrefinement}.

\section{Characterizing DREs}
\subsection{Definition and Taxonomy}

When answering a table-based question, LLMs are generally required to comprehend the table structure (e.g., distinguish between rows, understand column headers, and interpret each cell's meaning within its row-column context), use table values to support reasoning, and reason over critical ones to derive the correct answer.

Recent LLMs overcome long-standing challenges in tabular data by handling diverse textual formats with large-scale pre-training \citep{DBLP:journals/corr/abs-2307-09288} and enhancing logical and numerical reasoning through targeted post-training \citep{DBLP:journals/corr/abs-2503-20783,DBLP:journals/corr/abs-2506-20512}. However, our analysis reveals that these models, especially smaller ones, still make basic mistakes, which resemble human oversights that could have been avoided with careful attention. As shown in Figure~\ref{fig:intro}, the model confuses columns or overlooks an entire row. In CoT responses, we define \textbf{\emph{data referencing}} as the ability to correctly locate and cite information from inputs. Accordingly, errors or hallucinations in this process constitute \emph{data referencing errors} (DREs).

While DREs can be found in different domains and modalities \citep{DBLP:conf/iclr/MirzadehASTBF25,DBLP:conf/acl/HuangFMFFYZGWWH25}, in this work, we focus specifically on table-related tasks. Tables are highly data-intensive and often contain many similar rows and columns \citep{DBLP:journals/corr/abs-2501-19378}, which makes models particularly prone to referencing incorrect data. Formally, we categorize tabular DREs based on granularity of referenced content as follows:
\vspace{-2mm}
\begin{itemize*}
    \item \textbf{Incorrect Citation}: The response cites \textit{individual} table content (e.g., values or metadata) that does not match the actual table. This includes citing the wrong value, confusing rows or columns, or fabricating table-based content. As illustrated in Figure~\ref{fig:intro-1}, the model mistakenly took ``Nikkan Sports Grand Prix (Fall)'' as from the ``Award'' column, whereas the correct value should have been ``Best Supporting Actress''. This mix-up led to an incorrect final answer.
    \item \textbf{Omitted Information}: The response omits table values that belong to \textit{a required subset of the table}, such as listing all rows or identifying ``all teams with more than 5 wins.'' As shown in Figure~\ref{fig:intro-2}, the model correctly listed every row but missed the single row ``Oct 23''. This suggests that while the model can parse the table format, it still makes avoidable omissions.
\end{itemize*}
\vspace{-2mm}
In this work, we systematically investigate the occurrence and impact of DREs and propose a plug-in critic module to mitigate them. 

\begin{table*}[th]
    \small
    \centering
    \begin{tabular}{lcccc}
        \toprule
         & Accuracy (\%) & DRE Rate (\%) & DRE-in-Incorrect (\%) & Correct-in-DRE (\%) \\
        \midrule
        \multicolumn{5}{l}{\textbf{\textit{Qwen3-8B on different datasets}}} \\
        WTQ & 77.14 & 14.04 & 32.63 & 46.89 \\
        WTQ + prompting & 77.51 & 12.50 & 28.76 & 48.25 \\
        WTQ (CSV) & 75.94 & 17.54 & 37.32 & 48.82 \\
        WTQ (Markdown) & 77.26 & 14.34 & 32.69 & 48.15 \\
        TableBench & 77.48 & 10.55 & 30.63 & 13.43 \\
        FinQA & 63.21 & 33.57 & 39.34 & 56.88 \\
        SciTab & 77.53 & 14.06 & 21.54 & 65.57 \\
        ToTTo* & 14.06 & 18.45 & -- & -- \\
        \midrule
        \multicolumn{5}{l}{\textbf{\textit{Different models on WTQ}}} \\
        Qwen3-1.7B & 57.76 & 35.52 & 56.35 & 32.99 \\
        Qwen3-4B & 75.69 & 16.18 & 35.51 & 46.66 \\
        Qwen3-8B & 77.14 & 14.04 & 32.63 & 46.89 \\
        Qwen2.5-7B-Instruct & 43.32 & 17.56 & 23.60 & 23.85 \\
        Table-R1-Zero-7B & 76.10 & 19.29 & 52.41 & 35.08 \\
        Distill-Qwen-7B & 49.47 & 46.04 & 66.61 & 26.90 \\
        Distill-Llama-8B & 59.78 & 37.96 & 60.45 & 35.96 \\
        Llama4-Scout & 55.71 & 46.48 & 72.77 & 30.66 \\
        gpt-oss-20b & 78.38 & 5.71 & 16.29 & 38.31 \\
        \bottomrule
    \end{tabular}
    \caption{\textbf{DRE Evaluation Results} judged by Sonnet-3.7+gt. *: No binary correctness labels for ToTTo.}
    \vspace{-2mm}
    \label{tab:dre}
\end{table*}

\subsection{Evaluation via LLM-as-a-Judge}
\label{sec:evaluation}
To reduce human effort and enable automatic evaluation of DREs, we adopt LLM-as-a-Judge~\citep{DBLP:conf/nips/ZhengC00WZL0LXZ23,DBLP:journals/corr/abs-2505-07453}, leveraging a powerful LLM (\ie Sonnet-3.7, \citealp{anthropic2025claude37}) to detect DREs in model responses. To match human-level annotation quality, we address the following challenges by careful designs:
\begin{enumerate*}
    \item \textbf{Long\&Verbose Response:} Recent reasoning models often generate lengthy thinking processes \citep{DBLP:journals/tmlr/SuiCWZZYLWZZCH25}. To cope with this issue, we split the response at each occurrence of reflection tokens (\eg ``Wait'') and let the judge model evaluate one segment at a time. 
    \item \textbf{Detection Reliability:} Even strong models like Sonnet-3.7 can be swayed by the given response and fail to identify DREs, leading to false negatives (see Figure~\ref{fig:gt_need1},~\ref{fig:gt_need2}). To counter this, we provide the ground truth to the table-based question in the judge prompt. This helps the judge, especially when the final answer is wrong, to cross-check against the table more carefully and decide whether the error comes from a DRE.
\end{enumerate*}
In practice, the judge model is instructed to check whether a model-generated response uses table information accurately by examining the aforementioned two types of DREs---Incorrect Citations and Omitted Information. Manual inspection indicates that Sonnet-3.7 with ground truth (\ie Sonnet-3.7+gt) achieves an accuracy of 92.67\% with high consistency. Details and the full judge prompt are provided in Appendix~\ref{sec:app-llm-judge}. 

\paragraph{Evaluation Metrics} To holistically evaluate the occurrence of DREs in model responses to table questions, we calculate the following metrics:
\[
\text{DRE Rate} = \frac{|\text{DRE}|}{|\text{Total}|},
\]
where $|\text{DRE}|$ is the number of model responses containing at least one DRE. This metric measures the overall frequency of DREs.
\[
\text{Correct-in-DRE Ratio} = \frac{|\text{Correct} \cap \text{DRE}|}{|\text{DRE}|},
\]
where $|\text{Correct} \cap \text{DRE}|$ is the number of responses whose final answer is correct despite containing DREs. This metric captures DREs that cannot be detected by evaluating final-answer accuracy alone.
\[
\text{DRE-in-Incorrect Ratio} = \frac{|\text{Incorrect} \cap \text{DRE}|}{|\text{Incorrect}|},
\]
which provides an approximation of the correlation between DREs and final answer accuracy.

\subsection{Prevalence and Analysis}

Now, we examine the severity of DREs. We focus on three types of table tasks: \textbf{Question Answering}, including WTQ~\citep{DBLP:conf/acl/PasupatL15}, TableBench~\citep{DBLP:conf/aaai/WuYCZLDLSCSLLN25}, and FinQA~\citep{DBLP:conf/emnlp/ChenCSSBLMBHRW21}; \textbf{Claim Verification}, represented by SciTab~\citep{DBLP:conf/emnlp/LuPLNK23}, where the model is asked to determine whether a given claim is supported by the table; and \textbf{Table-to-Text}, represented by ToTTo~\citep{DBLP:conf/emnlp/ParikhWGFDYD20}, which requires generating a textual description conditioned on the table.

We evaluate a range of popular LLMs, spanning sizes from 1.7B to 20B and covering different model families: reasoning models that characterize extended thinking processes, such as Qwen3-8B \citep{DBLP:journals/corr/abs-2505-09388}; mixture-of-experts (MoE) models such as Llama4-Scout \citep{meta2025llama4}; and standard LLMs such as Qwen2.5-7B-Instruct \citep{DBLP:journals/corr/abs-2412-15115}. Following \citet{DBLP:conf/aaai/WuYCZLDLSCSLLN25}, we present tables in the JSON format, but we also experiment with CSV and Markdown formats. We further test a prompting-based method that explicitly instructs the model: \textit{Use only the table. Do not omit, miscite, or fabricate information. Ensure all cited values exactly match the table.} Model responses are then evaluated using Sonnet-3.7+gt, and the results are summarized in Table~\ref{tab:dre}. We have the following observations:

\textbf{(1) Data referencing errors are prevalent across different models, table formats, and table-related tasks.} For models, we observe that within a single model family such as Qwen3, data referencing capability improves with model size: larger models tend to produce fewer DREs. However, across different model families, this trend does not necessarily hold, as overall model capability also matters. For example, Llama4-Scout, as a non-reasoning model, shows relatively high rates of DREs (46.48\%) despite its size. Additionally, results across different table formats (JSON, CSV, and Markdown) and table-related tasks demonstrate that DREs cannot be attributed to specific formats or tasks, but instead represent a general and widespread challenge.

\textbf{(2) DREs persist under common mitigation strategies.} First, reasoning models including Qwen3 series, Distill series, and gpt-oss-20b are featured by self-reflection \citep{DBLP:journals/corr/abs-2501-12948,DBLP:journals/corr/abs-2408-03314}, yet they still exhibit DREs (5.71\%-46.04\%); in fact, once the first error is made, the model often repeats it, relying more on its own generation than on the original table (see Appendix Figure~\ref{fig:case-self-reflection} for an example). Second, even explicitly prompting the model to focus on data referencing accuracy, \ie WTQ + prompt setting, does not resolve DREs or improve final-answer accuracy. Third, Table-R1-Zero-7B \citep{DBLP:journals/corr/abs-2505-23621} was trained on table-related datasets from Qwen2.5-7B-Instruct using RLVR \citep{DBLP:journals/corr/abs-2411-15124}. While this specialized training improves answer accuracy, it does not effectively translate into fewer DREs, highlighting that data referencing is a separate capability that warrants further attention.

\textbf{(3) DREs may also occur during the reasoning process, even when the final answer is correct.} The Correct-in-DRE Ratio captures cases where the response contains DREs but still arrives at the correct final answer. This means that final-answer accuracy alone cannot guarantee the correctness of intermediate steps and the overall quality of the response. Besides, the Correct-in-DRE Ratio visibly varies across tasks. For example, SciTab shows a relatively high ratio (65.57\%), because its answers are binary labels (True or False). In such cases, numerical citation errors in reasoning process may not affect the final judgment, as illustrated in Figure~\ref{fig:case-scitab}.

\section{Reducing DREs with Critics}

We observe that DREs do occur in incorrect cases and can negatively impact final accuracy, as shown quantitatively in Table~\ref{tab:dre} and qualitatively in Figure~\ref{fig:intro}. Nevertheless, the DRE-in-Incorrect ratio should not be interpreted as indicating that this portion of incorrect answers is directly caused by DREs. This raises an important question: \textit{to what extent do DREs actually harm final accuracy?} In this section, we apply Sonnet-3.7+gt\footnote{Although provided with ground truth answers, Sonnet-3.7 does not directly judge final-answer correctness (see Appendix~\ref{sec:app-llm-judge}). We use Sonnet-3.7+gt to approximate the upper bound of a DRE detection critic.} as a high-quality critic to reduce DREs and explore whether this reduction translates into improvements in final-answer accuracy. We focus on three question answering datasets (WTQ, TableBench, FinQA) and three representative models (Qwen3-8B, Distill-Qwen-7B, Llama4-Scout).

\begin{table*}[t]
\small
\centering
\setlength{\tabcolsep}{6pt}
\begin{tabular}{lccccc}
\toprule
Dataset & Avg Acc (\%) & CF Acc (\%) & MV Acc (\%) & CF + MV Acc (\%) & \# Total \\
\midrule
\multicolumn{6}{l}{\textit{\textbf{Qwen3-8B}}} \\
WTQ  & 64.59 & 70.44 & 70.84 & \textbf{73.49} & 1509 \\
TableBench  & 63.12 & 67.42 & 70.17 & \textbf{71.82} & 181 \\
FinQA & 54.58 & 56.48 & 56.92 & \textbf{57.54} & 325 \\
\midrule
\multicolumn{6}{l}{\textit{\textbf{Distill-Qwen-7B}}} \\
WTQ & 49.47 & 61.83 & 62.05 & \textbf{65.80} & 2851 \\
TableBench & 55.06 & 69.62 & 67.60 & \textbf{71.65} & 321 \\
FinQA & 41.12 & 46.37 & 46.82 & \textbf{48.16} & 598 \\
\midrule
\multicolumn{6}{l}{\textit{\textbf{Llama4-Scout}}} \\
WTQ & 57.02 & 69.89 & 64.06 & \textbf{73.11} & 2265 \\
TableBench & 50.67 & 58.93 & 57.40 & \textbf{63.23} & 223 \\
FinQA & 39.53 & 42.32 & 44.91 & \textbf{46.76} & 216 \\
\bottomrule
\end{tabular}
\caption{Critic-based Filtering (CF) Results on the DRE subset. Avg Acc denotes the average accuracy over $N=8$ sampled responses per question. MV denotes Majority Voting, and CF + MV denotes majority-voting on critic-filtered subset.}
\vspace{-2mm}
\label{tab:cf-dre}
\end{table*}

\subsection{Critic-Based Filtering}

\paragraph{Method} A common application of a critic model is the Best-of-N (BoN) strategy, where an LLM generates multiple candidate responses and the critic selects the best one as the final output \citep{DBLP:journals/corr/abs-2408-03314,DBLP:journals/corr/abs-2212-08073,DBLP:journals/corr/abs-2307-09288}. While effective in some contexts, this approach assumes that the critic is able to fully judge the correctness of each response \citep{DBLP:journals/corr/abs-2110-14168} or assign highly discriminative scores across the set of responses \citep{DBLP:conf/iclr/LightmanKBEBLLS24,DBLP:journals/corr/abs-2211-14275}. Our critic, however, is designed specifically to detect DREs and thus cannot directly determine which single response is best overall. For example, multiple responses may contain no DREs yet still produce different final answers if mistakes occur later in the reasoning stage after retrieving the correct table values.

To address this limitation, we adopt a \textit{critic-based filtering} approach. Specifically, for a generation model (e.g., Qwen3-8B), we sample $N=8$ responses per question and use the critic to select the subset of responses with the fewest data referencing errors instead of selecting only one ``best'' response. This design improves the overall quality of the candidate pool and enables inference-time strategies such as majority voting to operate on a higher-quality set of responses, thereby further improving final accuracy.

\paragraph{Metrics} We report the average accuracy of all generated responses versus that of the subset selected by the critic. We also compare these results with majority voting. Our primary focus is the \emph{DRE subset}, which includes questions for which at least one response contains a data referencing error and at least one does not. This subset emphasizes data-referencing-challenging cases. To illustrate: if all sampled responses are free of referencing errors, then critic-based filtering will naturally show little or no improvement. Conversely, if all responses contain referencing errors, no selection strategy can guarantee correctness. Results on the full evaluation set are also reported in Appendix Table~\ref{tab:cf-full} but may underestimate the critic’s impact.

\paragraph{Results} From Table~\ref{tab:cf-dre}, we can observe that critic-based filtering steadily outperforms the average accuracy of all sampled responses by selecting those with fewer data referencing errors. This indicates that reducing DREs not only improves the quality of intermediate reasoning with fewer hallucinations but also translates into higher final accuracy. More encouragingly, it complements majority voting as an inference-time strategy: applying majority voting within the critic-filtered subset achieves the best performance, consistently surpassing majority voting alone. In some cases like Llama4-Scout on WTQ, even randomly selecting a response from the critic-filtered subset yields higher average accuracy than majority voting.

\subsection{Rejection Sampling}
\label{sec:rs}

\paragraph{Method} Another application of the critic model is rejection sampling. In standard LLMs, rejection sampling resembles BoN \citep{DBLP:journals/corr/abs-2212-08073,DBLP:journals/corr/abs-2307-09288}. In the context of reasoning models, this approach can be inefficient, as the responses of reasoning models are often very long \citep{DBLP:journals/corr/abs-2412-21187,DBLP:journals/tmlr/SuiCWZZYLWZZCH25} and thus costly to generate when sampling $N$ full completions. Moreover, repeated sampling increases computational expense.

We adapt rejection sampling for reasoning models by working at the segment level. Similar to Section~\ref{sec:evaluation}, for a generation model such as Qwen3-8B, we split a response into segments using the delimiter ``Wait''. Instead of regenerating an entire response, we selectively resample only the segment (or, when necessary, the entire response) until it passes the critic or reaches a maximum retry limit $N=8$ is reached. The model then continues with the next segment, repeating this process until the final answer is produced. This design reduces the cost of rejection sampling while preventing error propagation across the reasoning process.

\paragraph{Metrics} We report accuracy using rejection sampling on the DRE subset and the full set.

\begin{table}[t]
\small
\centering
\begin{tabular}{lll}
\toprule
Dataset & Acc in DRE (\%) & Acc in Full (\%) \\
\midrule
\multicolumn{3}{l}{\textit{\textbf{Qwen3-8B}}} \\
WTQ & 63.88 & 77.14 \\
\rowcolor{gray!15}
\quad + RS & 68.46{\scriptsize \textcolor{Green}{\,(+4.58)}} & 78.94{\scriptsize \textcolor{Green}{\,(+1.80)}} \\
TableBench & 63.54 & 77.48 \\
\rowcolor{gray!15}
\quad + RS & 69.09{\scriptsize \textcolor{Green}{\,(+5.55)}} & 79.31{\scriptsize \textcolor{Green}{\,(+1.83)}} \\
FinQA & 53.85 & 63.21 \\
\rowcolor{gray!15}
\quad + RS & 54.46{\scriptsize \textcolor{Green}{\,(+0.61)}} & 63.64{\scriptsize \textcolor{Green}{\,(+0.43)}} \\
\midrule
\multicolumn{3}{l}{\textit{\textbf{Distill-Qwen-7B}}} \\
WTQ & 48.58 & 49.47 \\
\rowcolor{gray!15}
\quad + RS & 57.14{\scriptsize \textcolor{Green}{\,(+8.56)}} & 55.99{\scriptsize \textcolor{Green}{\,(+6.52)}} \\
TableBench & 53.58 & 54.77 \\
\rowcolor{gray!15}
\quad + RS & 68.85{\scriptsize \textcolor{Green}{\,(+15.27)}} & 66.73{\scriptsize \textcolor{Green}{\,(+11.96)}} \\
FinQA & 41.14 & 46.90 \\
\rowcolor{gray!15}
\quad + RS & 44.82{\scriptsize \textcolor{Green}{\,(+3.68)}} & 49.52{\scriptsize \textcolor{Green}{\,(+2.62)}} \\
\midrule
\multicolumn{3}{l}{\textit{\textbf{Llama4-Scout}}} \\
WTQ & 56.16 & 55.71 \\
\rowcolor{gray!15}
\quad + RS & 65.39{\scriptsize \textcolor{Green}{\,(+9.23)}} & 61.92{\scriptsize \textcolor{Green}{\,(+6.21)}} \\
TableBench & 52.47 & 54.56 \\
\rowcolor{gray!15}
\quad + RS & 56.95{\scriptsize \textcolor{Green}{\,(+4.48)}} & 57.20{\scriptsize \textcolor{Green}{\,(+2.64)}} \\
FinQA & 40.74 & 58.94 \\
\rowcolor{gray!15}
\quad + RS & 43.52{\scriptsize \textcolor{Green}{\,(+2.78)}} & 59.72{\scriptsize \textcolor{Green}{\,(+0.78)}} \\
\bottomrule
\end{tabular}
\caption{Rejection Sampling Results. ``Acc in DRE'' denotes the results on the DRE subset, which we use as the primary evaluation setting.}
\vspace{-2mm}
\label{tab:rs}
\end{table}

\paragraph{Results} As shown in Table~\ref{tab:rs}, rejection sampling with the critic effectively improves final accuracy for both reasoning and non-reasoning models. As expected, the improvement is larger on the DRE subset than on the full set, since data referencing errors are more likely to appear in the DRE subset.
It is important to note that the rejection sampling process does not alter the generation model itself. A DRE-free response can be obtained by simply resampling. This highlights that DREs are largely avoidable errors rather than fundamental limitations of the model's knowledge or reasoning ability. However, how to reliably reduce the frequency of DREs remains an open problem, and rejection sampling with a DRE detection critic provides a promising and practical solution.

\section{Training a Small-Scale Critic}
In practice, ground-truth answers are often unavailable to the critic particularly during inference, and Sonnet-3.7 is both a black-box and costly to use. Therefore, in this section, we explore the feasibility of training a smaller-scale LLM (e.g. Qwen3-4B-Instruct) to perform the critic task.

\subsection{Small Critic Training}

\begin{figure*}
    \centering
    \includegraphics[width=\linewidth]{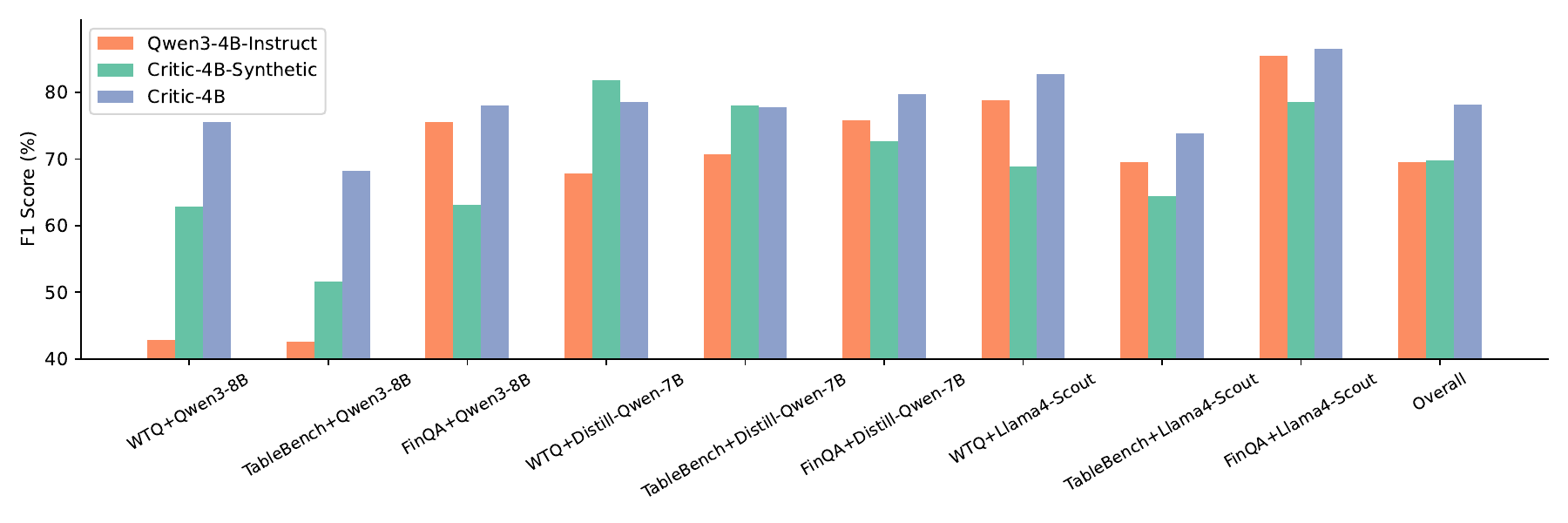}
    \vspace{-8mm}
    \caption{\small F1 scores across different model-dataset pairs for critic evaluation.}
    \label{fig:critic_evaluation}
    \vspace{-2mm}
\end{figure*}

For the critic model, similar to Section~\ref{sec:evaluation}, its task is as follows: given a table, a question based on the table, and a model’s response segment, determine whether the response contains DREs (See Appendix Figure~\ref{fig:critic-prompt} for the complete critic prompt). The output is either \textit{True} (contains DREs, treated as positive samples) or \textit{False} (does not contain DREs, including cases where no table values are cited, treated as negative samples). Beyond directly using Qwen3-4B-Instruct, we introduce a two-stage training pipeline to further enhance its ability to detect DREs.

\paragraph{(1) SFT.} We begin with supervised fine-tuning (SFT) as a warm-up stage to adapt Qwen3-4B-Instruct more effectively to the critic task. Although it already demonstrates strong instruction-following ability, we observe that training with RL directly, that is, without an SFT warm-up, causes the model to produce malformed outputs such as repeated <judgment></judgment> tags. To solve this, we use judgments from Sonnet-3.7 as distillation data in the first stage. This not only teaches Qwen3-4B-Instruct the expected output format but also transfers potentially useful ``critic heuristics'' from a stronger model, thereby stabilizing subsequent RL training.

\paragraph{(2) RLVR.} In the second stage, we employ reinforcement learning to enhance the critic model's robustness and generalization. Building on SFT foundation, RL enables exploration beyond the limitations of fixed supervision. We apply Reinforcement Learning with Verified Reward (RLVR, \citealp{DBLP:journals/corr/abs-2411-15124,DBLP:journals/corr/abs-2501-12948}), which is well suited to our setting: it leverages verified binary labels as reward signals and does not require constraining the CoT. This allows the critic to better detect DREs across diverse tasks.

For training data, we build on Qwen3-8B’s responses to the WTQ training set. We use Sonnet-3.7 to label all response segments, yielding a balanced dataset of 2,000 positive and negative samples for SFT, and 5,712 samples for RL training. The critic model trained using this data is called Critic-4B. In addition, we construct synthetic positives by inserting four types of DREs with rule-based heuristics (Appendix~\ref{sec:app-synthetic}), which reduces reliance on larger models and improves efficiency in both speed and cost. The critic model trained using the synthetic data is called Critic-4B-Synthetic.

\subsection{Critic Evaluation}

To evaluate a critic’s performance on the DRE detection task, we construct a \textit{critic evaluation dataset}. In details, we collect real positive and negative response segments, judged by Sonnet-3.7+gt, from the three models (Qwen3-8B, Distill-Qwen-7B, Llama4-Scout) across the three datasets (WTQ, TableBench, FinQA). This setup allows us to cover both reasoning (Qwen3-8B, Distill-Qwen-7B) and non-reasoning models (Llama4-Scout), as well as a diverse range of table-related question answering tasks spanning general-domain benchmarks and financial reasoning. For each model-dataset pair, we randomly sample 400 response segment with a balanced number of positive and negative examples, resulting in a total of 3,600 samples. We report the standard F1 score for this binary classification task, which balances precision (how often predicted DREs are correct) and recall (how many true DREs are identified).

\paragraph{Results} We compare the critic performance of Qwen3-4B-Instruct, Critic-4B-Synthetic, Critic-4B. The evaluation results are presented in Figure~\ref{fig:critic_evaluation}. We have the following findings:

Critic-4B consistently outperforms the untrained baseline Qwen3-4B-Instruct across all scenarios, achieving 78.16\% overall F1 compared to 69.51\%. Although it is trained only on the responses of Qwen3-8B on the WTQ training set, Critic-4B generalizes well to the same model’s responses on other table-related tasks. This is particularly notable because TableBench and FinQA's tables differ significantly from those in WTQ. In addition, Critic-4B achieves high F1 in identifying DREs in the responses of other models as well.

However, Critic-4B-Synthetic, trained on synthetic data, shows larger gains on in-distribution data, \ie on the same model or the same dataset. Yet, for settings with larger differences, such as FinQA (a different domain) and Llama4-Scout (non-reasoning model), critic performance actually declines. This suggests that the model may have overfit to biases specific to the synthetic data rather than learning to generalize to real-world errors.

\subsection{Rejection Sampling}
We also examine whether our trained small-scale critic, Critic-4B, can assist inference, using rejection sampling as described in Section~\ref{sec:rs}. We report both accuracy and DRE rate (on the full set) with and without rejection sampling.

\begin{table}[t]
\small
\centering
\resizebox{\linewidth}{!}{
\begin{tabular}{llll}
\toprule
Dataset & Acc in DRE $\uparrow$ & Acc in Full $\uparrow$ & DRE Rate $\downarrow$ \\
\midrule
\multicolumn{4}{l}{\textit{\textbf{Qwen3-8B}}} \\
WTQ & 63.88 & 77.14 & 14.04 \\
\rowcolor{gray!15}
\quad + RS & 66.37{\scriptsize \textcolor{Green}{\,(+2.49)}} & 
             78.25{\scriptsize \textcolor{Green}{\,(+1.11)}} & 
             10.08{\scriptsize \textcolor{Green}{\,(-3.96)}} \\
TableBench & 63.54 & 77.48 & 10.55 \\
\rowcolor{gray!15}
\quad + RS & 66.85{\scriptsize \textcolor{Green}{\,(+3.31)}} & 
             78.50{\scriptsize \textcolor{Green}{\,(+1.02)}} & 
             9.33{\scriptsize \textcolor{Green}{\,(-1.22)}} \\
FinQA & 53.85 & 63.21 & 33.57 \\
\rowcolor{gray!15}
\quad + RS & 54.46{\scriptsize \textcolor{Green}{\,(+0.61)}} & 
             63.47{\scriptsize \textcolor{Green}{\,(+0.26)}} & 
             31.21{\scriptsize \textcolor{Green}{\,(-2.36)}} \\
\midrule
\multicolumn{4}{l}{\textit{\textbf{Distill-Qwen-7B}}} \\
WTQ & 48.58 & 49.47 & 46.04 \\
\rowcolor{gray!15}
\quad + RS & 55.00{\scriptsize \textcolor{Green}{\,(+6.42)}} &
             54.01{\scriptsize \textcolor{Green}{\,(+4.54)}} &
             30.59{\scriptsize \textcolor{Green}{\,(-15.45)}} \\
TableBench & 53.58 & 54.77 & 46.86 \\
\rowcolor{gray!15}
\quad + RS & 63.24{\scriptsize \textcolor{Green}{\,(+9.66)}} & 
             61.66{\scriptsize \textcolor{Green}{\,(+6.89)}} & 
             32.45{\scriptsize \textcolor{Green}{\,(-14.41)}} \\
FinQA & 41.14 & 46.90 & 43.94 \\
\rowcolor{gray!15}
\quad + RS & 45.82{\scriptsize \textcolor{Green}{\,(+4.68)}} & 
             50.04{\scriptsize \textcolor{Green}{\,(+3.14)}} & 
             34.52{\scriptsize \textcolor{Green}{\,(-9.42)}} \\
\midrule
\multicolumn{4}{l}{\textit{\textbf{Llama4-Scout}}} \\
WTQ & 56.16 & 55.71 & 46.48 \\
\rowcolor{gray!15}
\quad + RS & 60.75{\scriptsize \textcolor{Green}{\,(+4.59)}} &
             58.45{\scriptsize \textcolor{Green}{\,(+2.74)}} &
             36.90{\scriptsize \textcolor{Green}{\,(-9.58)}} \\
TableBench & 52.47 & 54.56 & 38.54 \\
\rowcolor{gray!15}
\quad + RS & 55.16{\scriptsize \textcolor{Green}{\,(+2.69)}} & 
             56.39{\scriptsize \textcolor{Green}{\,(+1.83)}} & 
             33.87{\scriptsize \textcolor{Green}{\,(-4.67)}} \\
FinQA & 40.74 & 58.94 & 30.69 \\
\rowcolor{gray!15}
\quad + RS & 43.06{\scriptsize \textcolor{Green}{\,(+2.32)}} & 
             59.46{\scriptsize \textcolor{Green}{\,(+0.52)}} & 
             28.68{\scriptsize \textcolor{Green}{\,(-2.01)}} \\
\bottomrule
\end{tabular}}
\caption{\small Rejection Sampling Results (\%) using Critic-4B.}
\vspace{-4mm}
\label{tab:rs_ours}
\end{table}

As shown in Table~\ref{tab:rs_ours}, rejection sampling with Critic-4B consistently achieves higher accuracy compared to the setting without rejection sampling. Although it is less effective than the stronger critic, Sonnet-3.7+gt (refer to Table~\ref{tab:rs}), Critic-4B offers a lightweight and cost-effective alternative. Encouragingly, Critic-4B is smaller than all three generation models and does not benefit from the extended thinking processes as reasoning models like Qwen3-8B and Distill-Qwen-7B, yet it is still able to improve their accuracy. In addition, the DRE rate decreases when using rejection sampling, indicating that the critic not only enhances final-answer accuracy but also improves the overall quality of the model responses by reducing DREs.

\vspace{-1mm}
\section{Conclusions}
\vspace{-1mm}
We show that data referencing errors (DREs) are a pervasive weakness of LLMs on table reasoning tasks, undermining both response quality and final accuracy. By systematically analyzing DREs via LLM-as-a-Judge, we demonstrated their prevalence and propose inference-time strategies and lightweight critics to mitigate them. Our findings establish data referencing as a key evaluation dimension beyond final-answer accuracy for developing more reliable table reasoning systems.

\section*{Limitations}
Several limitations remain that warrant future study. First, we focus solely on DREs in table-related tasks, as they are a common and non-negligible issue. However, we recognize that DREs also arise in other domains and modalities. For instance, \citet{cobbe2021gsm8k,DBLP:conf/iclr/MirzadehASTBF25} describe the case: \textit{He makes 48 total ice cubes, including 10 giant cubes, 14 small cubes, 12 medium cubes, and some tiny cubes.} Qwen3-8B sometimes mistakenly interprets the order as giant, medium, small, tiny, which leads to errors. We hope future work can generalize to such broader domains.

Second, we did not examine the causes of DREs from an interpretability perspective. In preliminary experiments, we did observe that when the model \textit{prepared} to reference a table value, increasing its attention to the entire table helped reduce subsequent errors, suggesting that DREs are linked to insufficient attention. However, due to resource constraints, we did not scale up attention analyses or steering experiments. Future work could build on this direction.

\section*{Acknowledgments}

We are grateful to Robin Jia, He Wang, and Zelin He for their insightful feedback and discussions throughout this work. We also thank the anonymous reviewers for their constructive comments.

\bibliography{custom}

\newpage
\appendix

\section{LLM-as-a-Judge}
\label{sec:app-llm-judge}

The complete judge prompt for Sonnet-3.7+gt is shown in Figure~\ref{fig:sonnet37-gt}. Note that we provide ground-truth answers to mitigate false negatives, as explained in Section~\ref{sec:evaluation}. We also explicitly instruct the judge to \emph{focus solely on comparing the model response with the table data in order to assess table-referencing accuracy}. With this explicit instruction, we find that Sonnet-3.7+gt remains unbiased: it distinguishes reasoning mistakes from genuine DREs rather than assuming that every wrong final answer reflects a DRE. A case of Sonnet-3.7+gt’s judgment is shown below:

{\small \texttt{While the final answer differs from the reference answer, this appears to be a calculation error rather than a table referencing error. The model accurately extracted and cited all relevant values from the table.}}

We randomly sampled 100 instances from the critic evaluation dataset and three annotators at the PhD level independently assessed whether Sonnet-3.7+gt’s judgments were correct. Their assessments yielded an average accuracy of 92.67\%, indicating near-human reliability.

\section{Critic-based Filtering}
\label{sec:appendix}

\begin{table}[ht]
\small
\centering
\begin{tabular}{lccc}
\toprule
Dataset & Avg Acc (\%) & CF Acc (\%) & \# Total \\
\midrule
\multicolumn{3}{l}{\textit{\textbf{Qwen3-8B}}} \\
WTQ  & 77.55 & 79.65 & 4344 \\
TableBench  & 76.27 & 78.05 & 493 \\
FinQA & 64.01 & 64.74 & 1147 \\
\midrule
\multicolumn{3}{l}{\textit{\textbf{Distill-Qwen-7B}}} \\
WTQ & 49.60 & 58.43 & 4344 \\
TableBench & 55.17 & 65.87 & 493 \\
FinQA & 47.56 & 50.68 & 1147 \\
\midrule
\multicolumn{3}{l}{\textit{\textbf{Llama4-Scout}}} \\
WTQ & 56.22 & 62.93 & 4344 \\
TableBench & 53.80 & 57.54 & 493 \\
FinQA & 57.96 & 58.48 & 1147 \\
\bottomrule
\end{tabular}
\caption{Accuracy comparison between all-sample average and critic-selected subset on the full set.}
\label{tab:cf-full}
\end{table}

\section{Synthetic Positives Construction}
\label{sec:app-synthetic}
We use four strategies to insert DREs given a table and a model's response with correct final answers:
\begin{enumerate*}
    \item \textbf{Mix up rows:} Swap the identified value with a value from the same column but a different row.
    \item \textbf{Mix up columns:} Swap the value with another value from the same row but a different column.
    \item \textbf{Remove row:} Delete the entire row that contains the used value.
    \item \textbf{Remove a listed row:} Keep the table unchanged, but if the response enumerates all rows, randomly remove one row from the response and re-index.
\end{enumerate*}
We then use Qwen3-8B to perform inference for three times again to see whether the answer changes, and only save the cases where the final answer differs. For each saved case, this indicates that the modified table with the original model response, or the original table with the modified response, do not fully match---\ie there are DREs.

\section{Training Details}

For SFT, we use Llama-Factory \citep{zheng2024llamafactory} with a learning rate of 1e-5, a batch size of 8, 2,000 training examples, and train for 2 epochs. For RLVR, we use verl \citep{sheng2024hybridflow}, adopting GRPO \citep{DBLP:journals/corr/abs-2402-03300}, with a batch size of 256 and 8 rollouts per prompt at a temperature of 1.0. The learning rate is fixed at 1e-6, and we train for 20 epochs. During inference, we apply greedy decoding for the trained critic.

\section{Generation Details}
Except for ToTTO, we use string matching to compare accuracy across different table-related tasks. For ToTTO, following \citet{DBLP:journals/corr/abs-2505-23621}, we use $(\text{BLEU} + \text{ROUGE-L}) / 2$. For TableBench, we focus only on the Fact Checking and Numerical Reasoning subsets (493 in total), as the other two subsets, Data Analysis and Visualization, are beyond the tested models’ capabilities.

All generation models perform inference with their recommended decoding hyperparameters, as detailed in Table~\ref{tab:decoding}. For Llama4-Scout, we use the fp4 quantized version.\footnote{https://huggingface.co/nvidia/Llama-4-Scout-17B-16E-Instruct-FP4}

\begin{table}[ht]
    \small
    \centering
    \begin{tabular}{lccc}
        \toprule
        Model & Temperature & Top-P & Top-K \\
        \midrule
        Qwen3 Series & 0.6 & 0.95 & 20 \\
        Distill Series & 0.6 & 0.95 & -1 \\
        Llama4-Scout & 0.8 & 0.95 & -1 \\
        Qwen2.5-7B-Instruct & 0.6 & 0.95 & -1 \\
        Table-R1-Zero-7B & 0.6 & 0.96 & -1 \\
        gpt-oss-20b & 1.0 & 1.0 & -1 \\
        \bottomrule
    \end{tabular}
    \caption{Decoding hyperparamters used for generation models.}
    \label{tab:decoding}
\end{table}

\section{Code of Ethics}

All datasets and models we use are public. No ethical, safety, or privacy risks are involved in this study.

The licenses of the datasets and models we use are listed in Table~\ref{tab:license}.

\begin{table}[ht]
    \small
    \centering
    \resizebox{\linewidth}{!}{
    \begin{tabular}{lc}
        \toprule
         & License \\
        \midrule
        WTQ & CC-BY-SA-4.0 \\
        TableBench & Apache 2.0 \\
        FinQA & MIT \\
        SciTab & MIT \\
        ToTTo & Creative Commons Share-Alike 3.0 \\
        \midrule
        Qwen3 Series & Apache 2.0 \\
        Distill Series & MIT \\
        Llama4-Scout & llama4, nvidia-open-model-license \\
        Qwen2.5-7B-Instruct & Apache 2.0 \\
        Table-R1-Zero-7B & Apache 2.0 \\
        gpt-oss-20b & Apache 2.0 \\
        \bottomrule
    \end{tabular}}
    \caption{Licenses for datasets and models used in this paper.}
    \label{tab:license}
\end{table}

This paper used LLMs to polish writing. All original content came from the authors themselves.

\begin{figure*}[t]
\small
\centering
\begin{tcolorbox}[
  enhanced,
  title=Critic Prompt for Sonnet-3.7+gt,
  separator sign={\tcbline},
  separator sign dash={3pt}{3pt},
]

\textbf{\# Table Referencing Accuracy Evaluation}\\

\textbf{\#\# Task} \\
Your task is to evaluate whether a model-generated response accurately uses information from a given table. You will focus exclusively on table referencing accuracy, not other types of errors.\\

\textbf{\#\# Input Data} \\
<table>\{table\}</table>\\

<question>\{question\}</question>\\

<model\_response>\{model\_response\}</model\_response>\\

You are also provided with a reference answer, but this does not mean there is table referencing error: \\
<reference\_answer>\{answer\}</reference\_answer>\\

\textbf{\#\# Evaluation Instructions} \\
Analyze the model response sentence by sentence, focusing only on how accurately it references data from the table. If the response uses any value from the table, perform the following checks:\\

\textbf{\#\#\# Check 1: Copied Values Consistency} \\
- Identify any value that is explicitly copied or cited from the table. \\
- For each value: \\
\hspace*{2em}- Check if it exists in the table. \\
\hspace*{2em}- Check if it is used in the correct context (e.g., correct entity, row, or column). \\
- If a value is hallucinated, incorrect, or mismatched, it is a table referencing error.\\

\textbf{\#\#\# Check 2: Omission} \\
- When the model response tries to list all rows or items from the table, or items that meet a specific condition (e.g., ``all teams with more than 5 wins''), \\
- Then compare the listed items with the table. \\
- If any expected value is missing, it is a table referencing error.\\

\textbf{\#\# Important Guidelines} \\
- Focus ONLY on table referencing accuracy \\
- Ignore other types of errors (reasoning errors, calculation errors, question misunderstandings, etc.) \\
- The reference answer is provided for context only---base your judgment solely on comparing the model response with the table data\\

\textbf{\#\# Output Format} \\
After your analysis, provide your judgment using one of these formats:\\

1. If you find any table referencing error: \\
<judgment>Failed [Error Type] Check.</judgment> \\
Where [Error Type] is either ``Copied Values Consistency'' or ``Omission''\\

2. If no table referencing errors are found: \\
<judgment>NA</judgment>\\

Provide your judgment without additional explanation within the judgment tags.\\

Now follow the instructions step by step and put your final judgment within <judgment></judgment> tags.

\end{tcolorbox}
\vspace{-2mm}
\caption{Judge prompt for Sonnet-3.7+gt. Both ``Failed Copied Values Consistency Check'' and ``Failed Omission Check'' represents there are DREs.}
\label{fig:sonnet37-gt}
\end{figure*}

\begin{figure*}[t]
\small
\centering
\begin{tcolorbox}[
  enhanced,
  title=Failure of Self-Reflection in Reasoning Models,
  separator sign={\tcbline},
  separator sign dash={3pt}{3pt},
]
\textbf{Table}
\vspace{1ex}

\resizebox{\textwidth}{!}{
\begin{tabular}{|l|l|l|l|l|c|l|l|}
\hline
Date & Competition & Location & Country & Event & Placing & Rider & Nationality \\
\hline
31 October 2008 & 2008--09 World Cup & Manchester & United Kingdom & Sprint & 1 & Victoria Pendleton & GBR \\ \hline
31 October 2008 & 2008--09 World Cup & Manchester & United Kingdom & Keirin & 2 & Jason Kenny & GBR \\ \hline
1 November 2008 & 2008--09 World Cup & Manchester & United Kingdom & Sprint & 1 & Jason Kenny & GBR \\ \hline
1 November 2008 & 2008--09 World Cup & Manchester & United Kingdom & 500 m time trial & 1 & Victoria Pendleton & GBR \\ \hline
2 November 2008 & 2008--09 World Cup & Manchester & United Kingdom & Team sprint & 1 & Ross Edgar & GBR \\ \hline
2 November 2008 & 2008--09 World Cup & Manchester & United Kingdom & Team sprint & 1 & Jason Kenny & GBR \\ \hline
2 November 2008 & 2008--09 World Cup & Manchester & United Kingdom & Team sprint & 1 & Jamie Staff & GBR \\ \hline
2 November 2008 & 2008--09 World Cup & Manchester & United Kingdom & Keirin & 1 & Victoria Pendleton & GBR \\ \hline
2 November 2008 & 5th International Keirin Event & Manchester & United Kingdom & International keirin & 2 & Ross Edgar & GBR \\ \hline
13 February 2009 & 2008--09 World Cup & Copenhagen & Denmark & Team sprint & 1 & Chris Hoy & GBR \\ \hline
13 February 2009 & 2008--09 World Cup & Copenhagen & Denmark & Team sprint & 1 & Jason Kenny & GBR \\ \hline
13 February 2009 & 2008--09 World Cup & Copenhagen & Denmark & Team sprint & 1 & Jamie Staff & GBR \\ \hline
13 February 2009 & 2008--09 World Cup & Copenhagen & Denmark & Sprint & 1 & Victoria Pendleton & GBR \\ \hline
30 October 2009 & 2009--10 World Cup & Manchester & United Kingdom & Keirin & 1 & Chris Hoy & GBR \\ \hline
30 October 2009 & 2009--10 World Cup & Manchester & United Kingdom & Sprint & 1 & Victoria Pendleton & GBR \\ \hline
30 October 2009 & 2009--10 World Cup & Manchester & United Kingdom & Sprint & 1 & Chris Hoy & GBR \\ \hline
30 October 2009 & 2009--10 World Cup & Manchester & United Kingdom & 500 m time trial & 2 & Victoria Pendleton & GBR \\ \hline
1 November 2009 & 2009--10 World Cup & Manchester & United Kingdom & Team sprint & 1 & Ross Edgar & GBR \\ \hline
1 November 2009 & 2009--10 World Cup & Manchester & United Kingdom & Team sprint & 1 & Chris Hoy & GBR \\ \hline
1 November 2009 & 2009--10 World Cup & Manchester & United Kingdom & Team sprint & 1 & Jamie Staff & GBR \\
\hline
\end{tabular}}

\vspace{2ex}
\textbf{Question}

What is the number of 1st place finishes across all events?

\vspace{2ex}
\textbf{Model Response}

<think>\\
Okay, let's tackle this question. The user wants to know the number of 1st place finishes across all events from the provided table. \\

First, I need to look at the table data. The relevant columns here are ``Placing'' and ``Event''. Each row represents a rider's performance in a specific event. The ``Placing'' column indicates their position, so I need to count how many times ``1'' appears in that column.\\

Let me go through each row one by one:\\

Row 1: 1\textbackslash\textbackslash Row 2: 2\textbackslash\textbackslash Row 3: 1\textbackslash\textbackslash Row 4: 1\textbackslash\textbackslash Row 5: 1\textbackslash\textbackslash Row 6: 1\textbackslash\textbackslash Row 7: 1\textbackslash\textbackslash Row 8: 1\textbackslash\textbackslash Row 9: 2\textbackslash\textbackslash Row 10: 1\textbackslash\textbackslash Row 11: 1\textbackslash\textbackslash Row 12: 1\textbackslash\textbackslash Row 13: 1\textbackslash\textbackslash Row 14: 1\textbackslash\textbackslash Row 15: 1\textbackslash\textbackslash \highlight{red20}{Row 16: 2}\textbackslash\textbackslash \highlight{red20}{Row 17: 1}\textbackslash\textbackslash Row 18: 1\textbackslash\textbackslash Row 19: 1\\

So counting the 1s: rows 1,3,4,5,6,7,8,10,11,12,13,14,15,17,18,19. That's 16 instances. Wait, maybe I miscounted earlier. Let me list them:\\

Row 1: 1\textbackslash\textbackslash Row 3: 1\textbackslash\textbackslash Row 4: 1\textbackslash\textbackslash Row 5: 1\textbackslash\textbackslash Row 6: 1\textbackslash\textbackslash Row 7: 1\textbackslash\textbackslash Row 8: 1\textbackslash\textbackslash Row 10: 1\textbackslash\textbackslash Row 11: 1\textbackslash\textbackslash Row 12: 1\textbackslash\textbackslash Row 13: 1\textbackslash\textbackslash Row 14: 1\textbackslash\textbackslash Row 15: 1\textbackslash\textbackslash Row 17: 1\textbackslash\textbackslash Row 18: 1\textbackslash\textbackslash Row 19: 1\\

That's 16 entries where placing is 1. Wait, but \highlight{red20}{there are 19 rows total.} Let me check again. The rows are from 0 to 18 (assuming 19 rows). Let me check each row's placing:\\

Row 0: 1\textbackslash\textbackslash Row 1: 2\textbackslash\textbackslash Row 2: 1\textbackslash\textbackslash Row 3: 1\textbackslash\textbackslash Row 4: 1\textbackslash\textbackslash Row 5: 1\textbackslash\textbackslash Row 6: 1\textbackslash\textbackslash Row 7: 1\textbackslash\textbackslash Row 8: 2\textbackslash\textbackslash Row 9: 1\textbackslash\textbackslash Row 10: 1\textbackslash\textbackslash Row 11: 1\textbackslash\textbackslash Row 12: 1\textbackslash\textbackslash Row 13: 1\textbackslash\textbackslash Row 14: 1\textbackslash\textbackslash \highlight{red20}{Row 15: 2}\textbackslash\textbackslash \highlight{red20}{Row 16: 1}\textbackslash\textbackslash Row 17: 1\textbackslash\textbackslash Row 18: 1\\

So that's 1 (row0), 1 (row2), 1 (row3), 1 (row4), 1 (row5), 1 (row6), 1 (row7), 1 (row9), 1 (row10), 1 (row11), 1 (row12), 1 (row13), 1 (row14), 1 (row16), 1 (row17), 1 (row18). That's 16 instances. So the answer should be 16. \\

Wait, but let me check again. Let me count each row's placing:\\

Row 0: 1\textbackslash\textbackslash Row 1: 2\textbackslash\textbackslash Row 2: 1\textbackslash\textbackslash Row 3: 1\textbackslash\textbackslash Row 4: 1\textbackslash\textbackslash Row 5: 1\textbackslash\textbackslash Row 6: 1\textbackslash\textbackslash Row 7: 1\textbackslash\textbackslash Row 8: 2\textbackslash\textbackslash Row 9: 1\textbackslash\textbackslash Row 10: 1\textbackslash\textbackslash Row 11: 1\textbackslash\textbackslash Row 12: 1\textbackslash\textbackslash Row 13: 1\textbackslash\textbackslash Row 14: 1\textbackslash\textbackslash \highlight{red20}{Row 15: 2}\textbackslash\textbackslash \highlight{red20}{Row 16: 1}\textbackslash\textbackslash Row 17: 1\textbackslash\textbackslash Row 18: 1\\

Yes, that's 16 times. So the final answer is 16.\\
</think>\\

\highlight{gray30}{Final Answer: 16}\\
\end{tcolorbox}
\vspace{-2mm}
\caption{This is an example of Qwen3-8B on the WTQ test set. The table contains 20 rows in total, but Qwen3-8B identifies only 19, missing the actual 16th row (Placing: 1). Even after repeated checks and attempts with different numbering schemes (starting from 1 or from 0), it consistently reproduced this DRE, ultimately leading to an incorrect answer.}
\label{fig:case-self-reflection}
\end{figure*}

\begin{figure*}[t]
\small
\centering
\begin{tcolorbox}[
  enhanced,
  title=Input,
  separator sign={\tcbline},
  separator sign dash={3pt}{3pt},
]
\textbf{Table}
\vspace{1ex}

\resizebox{\textwidth}{!}{
\begin{tabular}{|l|l|c|c|c|c|c|c|c|c|}
\hline
Setting & Metrics & cs-en & de-en & fi-en & lv-en & ru-en & \cellcolor{blue20} tr-en & zh-en & Avg \\
\hline
Baselines & METEOR++ & 0.552 & 0.538 & 0.720 & 0.563 & 0.627 & 0.626 & 0.646 & 0.610 \\ \hline
Baselines & RUSE(*) & 0.624 & 0.644 & 0.750 & 0.697 & 0.673 & 0.716 & 0.691 & 0.685 \\ \hline
Baselines & \cellcolor{blue20} BERTScore-F1 & 0.670 & 0.686 & 0.820 & 0.710 & 0.729 & \cellcolor{green20} 0.714 & 0.704 & 0.719 \\ \hline
Sent-Mover & Smd + W2V & 0.438 & 0.505 & 0.540 & 0.442 & 0.514 & 0.456 & 0.494 & 0.484 \\ \hline
Sent-Mover & Smd + ELMO + PMeans & 0.569 & 0.558 & 0.732 & 0.525 & 0.581 & 0.620 & 0.584 & 0.595 \\ \hline
Sent-Mover & Smd + BERT + PMeans & 0.607 & 0.623 & 0.770 & 0.639 & 0.667 & 0.641 & 0.619 & 0.652 \\ \hline
Sent-Mover & Smd + BERT + MNLI + PMeans & 0.616 & 0.643 & 0.785 & 0.660 & 0.664 & 0.668 & 0.633 & 0.667 \\ \hline
Word-Mover & Wmd-1 + W2V & 0.392 & 0.463 & 0.558 & 0.463 & 0.456 & 0.485 & 0.481 & 0.471 \\ \hline
Word-Mover & Wmd-1 + ELMO + PMeans & 0.579 & 0.588 & 0.753 & 0.559 & 0.617 & 0.679 & 0.645 & 0.631 \\ \hline
Word-Mover & Wmd-1 + BERT + PMeans & 0.662 & 0.687 & 0.823 & 0.714 & 0.735 & 0.734 & 0.719 & 0.725 \\ \hline
Word-Mover & Wmd-1 + BERT + MNLI + PMeans & 0.670 & 0.708 & \textbf{0.835} & \textbf{0.746} & \textbf{0.738} & 0.762 & \textbf{0.744} & \textbf{0.743} \\ \hline
Word-Mover & Wmd-2 + BERT + MNLI + PMeans & \textbf{0.679} & \textbf{0.710} & 0.832 & 0.745 & 0.736 & \textbf{0.763} & 0.740 & \textbf{0.743} \\
\hline
\end{tabular}}

\vspace{2ex}
\textbf{Claim}

Table 1: In all language pairs, the best correlation is achieved by our word mover metrics that use a BERT pretrained on MNLI as the embedding generator and PMeans to aggregate the embeddings from different BERT layers, i.e., WMD-1/2+BERT+MNLI+PMeans.

\vspace{2ex}
\textbf{Question}

Is the above claim true or false?

\vspace{2ex}
\textbf{Model Response}

...\\
6. **tr-en (Turkish-English):**\\
\quad - Wmd-1: 0.744\\
\quad - Wmd-2: 0.763\\
\quad - BERTScore-F1 is \highlight{red20}{0.704}. So Wmd-2 is higher here.\\
\\
...
\end{tcolorbox}
\vspace{-2mm}
\caption{This is an example of Qwen3-8B on the SciTab test set. Qwen3-8B misquoted 0.714 as 0.704, but this does not affect the subsequent conclusion that Wmd-2 (0.763) is higher. Therefore, final-answer accuracy cannot fully reflect the presence of a DRE.}
\label{fig:case-scitab}
\end{figure*}

\begin{figure*}[t]
\small
\centering
\begin{tcolorbox}[
  enhanced,
  title=Critic Prompt,
  separator sign={\tcbline},
  separator sign dash={3pt}{3pt},
]

You are a careful evaluator. Your task is to check whether a model's response uses table information accurately.\\

<instructions>  \\
Please follow these steps:\\

1. Review the inputs:  \\
<table>\{table\}</table>  \\
<question>\{question\}</question>  \\
<model\_response>\{model\_response\}</model\_response>\\

2. Go through the <model\_response> sentence by sentence. If the response uses any value from the table, do the following checks:\\

<check\_1>  \\
\textbf{Copied Values Consistency Check:}  \\
- Find any value that is explicitly copied or cited from the table.  \\
- For each value:  \\
\hspace*{2em}- Check if it exists in the table.  \\
\hspace*{2em}- Check if it is used in the correct context (e.g., correct entity, row, or column).  \\
- If a value is hallucinated, incorrect, or mismatched, it is an error.  \\
</check\_1>\\

<check\_2>  \\
\textbf{Omission Check:}  \\
- If the segment tries to list all rows from the table, or all items that meet a specific condition (e.g., “all teams with more than 5 wins”),  \\
- Then compare the listed items with the table.  \\
- If any expected value is missing, it is an error.  \\
</check\_2>\\

3. If any error is found, stop the evaluation immediately and respond with:  \\
<judgment>True</judgment>\\

4. If no errors are found in the entire segment, respond with:  \\
<judgment>False</judgment>\\

Follow the instructions step by step and put your final judgment within <judgment> </judgment> tags.  \\
</instructions> 

\end{tcolorbox}
\vspace{-2mm}
\caption{Critic Prompt for small-scale LLMs. We do not provide ground-truth answers in this prompt.}
\label{fig:critic-prompt}
\end{figure*}

\begin{figure*}[t]
\small
\centering
\begin{tcolorbox}[
  enhanced,
  title=Judge Prompt Input,
  separator sign={\tcbline},
  separator sign dash={3pt}{3pt},
]
\textbf{Table}
\vspace{1ex}

\begin{tabular}{|c|l|c|c|c|c|}
\hline
Rank & Nation        & Gold & \cellcolor{blue20} Silver & Bronze & Total \\
\hline
1  & Soviet Union   & \cellcolor{red20} *7* & \cellcolor{green20} 3 & 6 & 16 \\ \hline
2  & Austria        & 4 & 3 & 4 & 11 \\ \hline
3  & Finland        & 3 & 3 & 1 & 7 \\ \hline
4  & Switzerland    & 3 & 2 & 1 & 6 \\ \hline
5  & Sweden         & 2 & 4 & 4 & 10 \\ \hline
6  & United States  & 2 & 3 & 2 & 7 \\ \hline
7  & Norway         & 2 & 1 & 1 & 4 \\ \hline
8  & Italy          & 1 & 2 & 0 & 3 \\ \hline
9  & Germany        & 1 & 0 & 1 & 2 \\ \hline
10 & Canada         & 0 & 1 & 2 & 3 \\
\hline
\end{tabular}

\vspace{2ex}
\textbf{Question}

Who had the most silver medals?

\vspace{2ex}
\textbf{Model Response}

... The question is about silver medals. Let me list all the \highlight{blue20}{silver} counts:\\
\\
\highlight{red20}{Soviet Union: 7 (assuming the *7* is 7)}\\
Austria: 3\\
Finland: 3\\
Switzerland: 2\\
Sweden: 4\\
United States: 3\\
Norway: 1\\
Italy: 2\\
Germany: 0\\
Canada: 1\\
\\
So the maximum here is 7, which is the Soviet Union. Therefore, the answer should be Soviet Union.
\end{tcolorbox}
\vspace{-2mm}
\caption{An input example for the judge prompt. In this case, the generation model correctly cites the ``Silver'' value, except for the first row, ``Soviet Union.''}
\label{fig:gt_need1}
\end{figure*}

\begin{figure*}[t]
\small
\centering
\begin{tcolorbox}[
  enhanced,
  title=Judge Output,
  separator sign={\tcbline},
  separator sign dash={3pt}{3pt},
]
\textbf{Sonnet-3.7 (without ground truth)}

I'll analyze the model's response sentence by sentence, checking for any errors in how it uses data from the table.\\
\\
\highlight{red20}{The model correctly identifies that Soviet Union has "*7*" in the Silver column (with asterisks)}, and acknowledges this unusual formatting. It then reasonably interprets this as 7 while being transparent about this assumption.\\
\\
When listing silver medal counts for all countries, the model states:\\
- Soviet Union: 7 (with clear explanation about the "*7*" notation)\\
- Austria: 3 \checkmark\\
- Finland: 3 \checkmark\\
- Switzerland: 2 \checkmark\\
- Sweden: 4 \checkmark\\
- United States: 3 \checkmark\\
- Norway: 1 \checkmark\\
- Italy: 2 \checkmark\\
- Germany: 0 \checkmark\\
- Canada: 1 \checkmark\\
\\
\highlight{red20}{All these values match exactly with the silver medal counts in the table.} The model provides a complete list of all countries' silver medal counts without any omissions.\\
\\
\highlight{red20}{The model's conclusion that Soviet Union has the most silver medals (interpreting "*7*" as 7) is consistent with the data presented in the table.}\\
\\
\highlight{gray30}{<judgment>NA</judgment>}
\tcbline
\textbf{Sonnet-3.7 with ground truth}

I'll carefully evaluate the model response for table referencing accuracy by checking how it uses values from the provided table.\\
\\
First, let me examine the actual silver medal counts in the table:\\
\highlight{green20}{- Soviet Union: 3 silver medals (not 7)}\\
- Austria: 3 silver medals\\
- Finland: 3 silver medals\\
- Switzerland: 2 silver medals\\
- Sweden: 4 silver medals\\
- United States: 3 silver medals\\
- Norway: 1 silver medal\\
- Italy: 2 silver medals\\
- Germany: 0 silver medals\\
- Canada: 1 silver medal\\
\\
The model response makes a critical table referencing error. It incorrectly states that the Soviet Union has 7 silver medals when looking at the row: ["1", "Soviet Union", "*7*", "3", "6", "16"]. \highlight{green20}{The model misinterpreted the data, as the "*7*" appears in the Gold column, not the Silver column. The Soviet Union actually has 3 silver medals according to the table.}\\
\\
The model lists all silver medal counts but assigns the wrong value (7) to the Soviet Union. Based on this incorrect value, it concludes that the Soviet Union has the most silver medals, when in reality Sweden has the most with 4.\\
\\
\highlight{blue20}{<judgment>Failed Copied Values Consistency Check.</judgment>}
\end{tcolorbox}
\vspace{-2mm}
\caption{The judge output of Sonnet-3.7 without and with ground truth. Without a ground-truth answer, Sonnet-3.7 is misled by the model response; with a ground-truth answer, it checks the consistency between the model response and the table more carefully.}
\label{fig:gt_need2}
\end{figure*}

\end{document}